\documentclass{article}

\usepackage{microtype}
\usepackage{graphicx}
\usepackage{subfigure}
\usepackage{booktabs} % for professional tables

\usepackage{hyperref}

\usepackage{amsfonts} 
\usepackage{amsmath}

\usepackage[accepted]{ai4science}

\icmltitlerunning{Target-aware molecular graph generation}

\begin{document}

\twocolumn[
\icmltitle{Target-aware Molecular Graph Generation}



\icmlsetsymbol{equal}{*}

\begin{icmlauthorlist}
\icmlauthor{Cheng Tan}{equal,westlake,zju}
\icmlauthor{Zhangyang Gao}{equal,westlake,zju}
\icmlauthor{Stan Z. Li}{westlake}
\end{icmlauthorlist}

\icmlaffiliation{westlake}{AI Research and Innovation Lab, Westlake University}
\icmlaffiliation{zju}{Zhejiang University}

\icmlcorrespondingauthor{Stan Z. Li}{Stan.ZQ.Li@westlake.edu.cn}
\icmlcorrespondingauthor{Cheng Tan}{tancheng@westlake.edu.cn}

\icmlkeywords{Machine Learning, ICML}

\vskip 0.3in
]



\printAffiliationsAndNotice{\icmlEqualContribution} % otherwise use the standard text.

\begin{abstract}
Generating molecules with desired biological activities has attracted growing attention in drug discovery. Previous molecular generation models are designed as chemocentric methods that hardly consider the drug-target interaction, limiting their practical applications. In this paper, we aim to generate molecular drugs in a target-aware manner that bridges biological activity and molecular design. To solve this problem, we compile a benchmark dataset from several publicly available datasets and build baselines in a unified framework. Building on the recent advantages of flow-based molecular generation models, we propose \textit{SiamFlow}, which forces the flow to fit the distribution of target sequence embeddings in latent space. Specifically, we employ an alignment loss and a uniform loss to bring target sequence embeddings and drug graph embeddings into agreements while avoiding collapse. Furthermore, we formulate the alignment into a one-to-many problem by learning spaces of target sequence embeddings. Experiments quantitatively show that our proposed method learns meaningful representations in the latent space toward the target-aware molecular graph generation and provides an alternative approach to bridge biology and chemistry in drug discovery.
\end{abstract}

\section{Introduction}
Drug discovery, which focuses on finding candidate molecules with desirable properties for therapeutic applications, is a long-period, expensively process with a high failure rate. The challenge primarily stems from the actuality that only a tiny fraction of the theoretical possible drug-like molecules may have practical effects. Specifically, the entire search space is as large as $10^{23} \sim 10^{60}$, while only $10^{8}$ of them are therapeutically relevant~\cite{polishchuk2013estimation,hert2009quantifying,cheng2021molecular}. In the face of such difficulty, traditional methods like high-throughput screening~\cite{hert2009quantifying} fail in terms of efficiency because of the large amount of resources required in producing minor hit compounds. One alternative is using computational methods~\cite{phatak2009high,paricharak2018data} such as virtual screening~\cite{bajorath2002integration,schneider2010virtual} to identify hit compounds from virtual libraries through similarity-based searches or molecular docking. Another alternative is automated de novo molecule design, such as inverse QSAR~\cite{schneider2016novo}, particle swarm optimization~\cite{winter2019efficient}, structure-based de novo design~\cite{schneider2013novo,button2019automated}, or genetic algorithms~\cite{bandholtz2012molecular}.

Recent deep generative models have demonstrated potentials to promote drug discovery by exploring huge chemical space in a data-driven manner. Various forms of variational autoencoder (VAE)~\cite{kingma2013auto,gomez2018automatic,kusner2017grammar,dai2018syntax,jin2018junction,simonovsky2018graphvae}, generative adversarial networks (GAN)~\cite{goodfellow2014generative,guimaraes2017objective,sanchez2017optimizing,prykhodko2019novo,mendez2020novo}, autoregressive (AR)~\cite{van2016pixel,popova2019molecularrnn,you2018graph}, and normalizing flow (NF)~\cite{dinh2014nice,dinh2016density_realnvp,madhawa2019graphnvp,honda2019graph,shi2019graphaf,zang2020moflow,luo2021graphdf} have been proposed to generate molecular SMILES or graphs. Though these approaches can generate valid and novel molecules to some extent, they remain inefficient because the generated candidate molecules need further screened against given targets. As the primary goal of these chemocentric methods is to generate drug-like molecules that satisfy specific properties, directly applying them in drug discovery requires extra efforts on predicting the binding affinities between candidate molecules and target proteins. 

\begin{figure}[ht]
    \centering
    \includegraphics[width=0.48\textwidth]{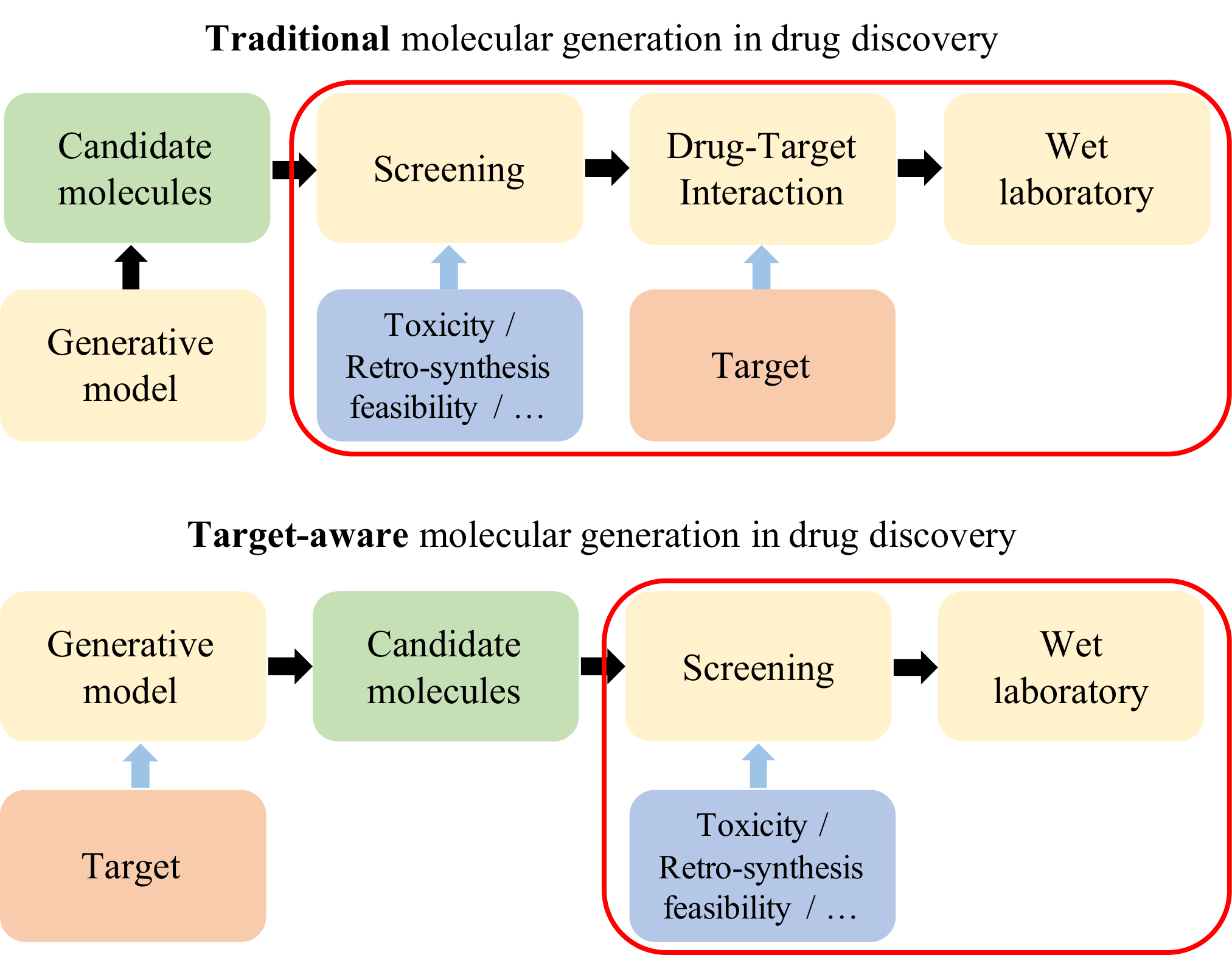}
    \caption{The computational drug discovery pipelines of traditional chemocentric and target-aware molecular generation. The black arrows denote the main steps, the blue arrows denote external considerations, and the red boxes denote the post-processing process of generated molecules in drug discovery.}
    \label{fig:comparison_tamg}
\end{figure}

While previous molecular generation methods scarcely take biological drug-target interactions into account, we aim to generate candidate molecules based on a biological perspective. This paper proposes target-aware molecular generation to bridge biological activity and chemical molecular design that generate valid molecules conditioned on specific targets and thus facilitate the development of drug discovery. 

As shown in Figure~\ref{fig:comparison_tamg}, the pipeline of computational drug discovery is supposed to be simplified to a great extent with the help of target-aware molecular generation.

Our main contributions are summarized as follows:

\begin{itemize}
\item We propose a target-aware molecular generation manner from a biological perspective, while prior works on chemocentric molecular generation are inefficient in practical drug discovery.
\item We establish a new benchmark for the target-aware molecular generation containing abundant drug-target pairs for evaluating generative models. 
\item We propose SiamFlow, a siamese network architecture for the conditional generation of flow-based models. While the sequence encoder and the generative flow align in the latent space, a uniformity regularization is imposed to avoid collapse. Moreover, we implement SiamFlow and baselines in a unified framework for further research. The code will be released.
\end{itemize}

\section{Related work}

\subsection{De Novo Molecular Generation}
Recent years have witnessed revolutionary advances in de novo molecular generation, which aims to generate molecules with desired properties from scratch. Artificial intelligence, which has been widely applied in this field, not only significantly reduces the chemical search space but also lowers time consumption to a large extent. 

\paragraph{VAE-based}

VAE has been attractive in molecular generation in the virtue of its latent space is potentially operatable. CharVAE~\cite{gomez2018automatic} first proposes to learn from molecular data in a data-driven manner and generate with a VAE model. By jointly training the task of predicting molecular properties from the hidden layers, this method explores the molecular space without manual prior knowledge. GVAE~\cite{kusner2017grammar} represents each data as a parse tree from a context-free grammar, and directly encodes to and decodes from these parse trees to ensure the validity of generated molecules. Inspired by syntax-directed translation in complier theory, SD-VAE~\cite{dai2018syntax} proposes to convert the offline syntax-directed translation check into on-the-fly generated guidance for ensuring both syntactical and semantical correctness. JT-VAE~\cite{jin2018junction} first realize the direct generation of molecular graphs instead of linear SMILES (Simplified Molecular-Input Line-Entry System) strings. HierVAE~\cite{jin2020hierarchical} proposes a motif-based hierarchical encoder-decoder toward the large molecular graph generation.

\paragraph{GAN-based}

An alternative is to implement GAN in molecular generation. ORGAN~\cite{guimaraes2017objective} adds expert-based rewards under the framework of WGAN~\cite{arjovsky2017wasserstein}. ORGANIC~\cite{sanchez2017optimizing} improves the above work for inverse design chemistry and implements the molecular generation towards specific properties. MolGAN~\cite{de2018molgan} proposes GAN-based models to generate molecular graphs rather than SMILES. Motivated by cycle-consistent GAN~\cite{CycleGAN2017}, Mol-CycleGAN~\cite{maziarka2020mol} generates optimized compounds with high structural similarity to the original ones.

\paragraph{Flow-based}

Molecular generation with the normalizing flow is promising as its invertible mapping can reconstruct the data exactly. GraphNVP~\cite{madhawa2019graphnvp} defines a normalizing flow from a base distribution to the molecular graph structures. GRF~\cite{honda2019graph} further explores invertible mappings by residual flows, and presents a way of keeping the entire flows invertible throughout the training and sampling process in molecular generation. GraphAF~\cite{shi2019graphaf} combines the advantages of both autoregressive and flow-based approaches to iteratively generate molecules. MolFlow~\cite{zang2020moflow} proposes a variant of Glow~\cite{kingma2018glow} to generate atoms and bonds in a one-shot manner. MolGrow~\cite{kuznetsov2021molgrow} constrains optimization of properties by using latent variables of the model, and recursively splits nodes to generate molecular structures.

% 应再加一段中总结的，前人...，但是...

Though these approaches have achieved significant performance, we recognize them as chemocentric molecular generation methods that lack biological connections. We aim to bridge biological and chemical perspectives in molecular generation for practical drug discovery.

\subsection{Drug-target Interaction}

Drug-target interaction (DTI) has been extensively developed over the decades. Computational virtual screening methods like molecular docking~\cite{trott2010autodock} and molecular dynamics simulations~\cite{salsbury2010molecular} have provided mechanistic insights in this field. Though these methods have inherently great interpretability, they suffer from heavy dependence on the available three-dimensional structure data with massive computational resources. 

Recent progress in artificial intelligence has inspired researchers to utilize deep learning techniques in drug-target interaction prediction. DeepDTA~\cite{ozturk2018deepdta} and DeepAffinity~\cite{karimi2019deepaffinity} are representatives of deep-learning-based methods that take SMILES of drugs and primary sequences of proteins as input, from which neural networks are employed to predict affinities. InterpretableDTIP~\cite{gao2018interpretable} predicts DTI directly from low-level representations and provides biological interpretation using a two-way attention mechanism. DeepRelations~\cite{karimi2020explainable} embeds protein sequences by hierarchical recurrent neural network and drug graphs by graph neural networks with joint attention between protein residues and compound atoms.
DEEPScreen~\cite{rifaioglu2020deepscreen} learns features from structural representations for a large-scale DTI prediction. MONN~\cite{li2020monn} predicts both pairwise non-covalent interactions and binding affinities between drugs and targets with extra supervision from the labels extracted from available high-quality three-dimensional structures.

% 总结

Our proposed target-aware molecular generation builds on the recent advances in data-driven drug-target interaction prediction. We connect chemical molecular generation with biological drug-target interaction to promote the efficiency of computational drug discovery.

\subsection{Conditional Molecular Generation}

Generating molecules with the consideration of some external conditions is a promising field. There is plenty of work that conditioning on desired properties based on chemical perspective. CVAE~\cite{gomez2018automatic} jointly trains VAE with a predictor that predicts properties from the latent representations of VAE. \cite{lim2018molecular} proposes applying conditional VAE to generate drug-like molecules satisfying five properties at the same time: MW (molecular weight), LogP (partition coefficient), HBD (number of hydrogen bond donor), HBA (number of hydrogen acceptor), and TPSA (topological polar surface area). \cite{griffiths2020constrained} employs constrained Bayesian optimization to control the latent space of VAE in order to find molecules that score highly under a specified objective function which is a weighted function of LogP and QED (quantitative estimate of drug-likeness). CogMol~\cite{chenthamarakshan2020cogmol} and CLaSS~\cite{das2021accelerated} pretrain the latent space with SMILES and train property classifiers from the latent representations. They sample variables from the latent space that satisfy high scores from property classifiers to generate molecules. 

Though recent molecular generation methods~\cite{jin2018junction,madhawa2019graphnvp,zang2020moflow,luo2021graphdf} also present property optimization experiments, they still, like the above methods, barely take account of drug-target interaction. \cite{mendez2020novo} proposes stacks of conditional GAN to generate hit-like molecules from gene expression signature. While this work focuses on drug-gene relationships, we recognize that drug-protein is a more typical case.

\section{Background and Preliminaries}

\subsection{Problem Statement} Let $\mathcal{T} = \{T_i\}_{i=1}^t$ be a set of targets, and there exists a set of drugs $\mathcal{M}_{T_i} = \{M_j^{(T_i)} \}_{j=1}^{d_i}$ that bind to each target $T_i$. $S(T, M)$ is defined as a function measuring the interaction between target $T$ and drug $M$. The target-aware molecular generation aims to learning a generation model $p_\theta(\cdot|T_i)$ from each drug-target pair $(M_j^{(T_i)}, T_i)$ so as to maximize $\mathbb{E}_{M|T_i \sim p_\theta}[S(M, T_i)]$. 

\subsection{The Flow Framework} A flow model is a sequence of parametric invertible mapping $f_\Theta = f_Q \circ ... \circ f_1$ from the data point $x \in \mathbb{R}^D$ to the latent variable $z \in \mathbb{R}^D$, where $x \sim P_X(x), z \sim P_Z(z)$. The latent distribution $P_Z$ is usually predefined as a simple distribution, e.g., a normal distribution. The complex data in the original space is modelled by using the change-of-variable formula:

\begin{equation}
    P_X(x) = P_Z(z) \bigg| \mathrm{det}\frac{\partial Z}{\partial X} \bigg|,
\end{equation}
and its log-likelihood:
\begin{equation}
\begin{aligned}
    \log P_X(x) &= \log P_Z(z) + \log \bigg| \mathrm{det}\frac{\partial Z}{\partial X} \bigg| \\
    &= \log P_Z(z) + \sum_{q=1}^Q \log \bigg| \mathrm{det}\frac{\partial f_q (z^{(q-1)})}{\partial z^{(q-1)}} \bigg| ,
\end{aligned}
\end{equation}
where $z^{(q)} = f_q(z^{(q-1)})$, and we represent the input $z^{(0)}$ by using $z$ for notation simplicity.

As the calculation of the Jacobian determinant for $f_\Theta$ is expensive for arbitrary functions, NICE~\cite{dinh2014nice} and RealNVP~\cite{dinh2016density_realnvp} develop an affine coupling transformation $z = f_\Theta(x)$ with expressive structures and efficient computation of the Jacobian determinant.

For given $D$-dimensional input $x$ and $d < D$, the output $y$ of an affine coupling transformation is defined as:
\begin{equation}
\begin{aligned}
    y_{1:d} &= x_{1:d} \\
    y_{d+1:D} &= x_{d+1:D} \odot \exp (S_\Theta(x_{1:d})) + T_\Theta(x_{1:d}),
\end{aligned}
\end{equation}
where $S_\Theta: \mathbb{R}^{d} \rightarrow \mathbb{R}^{D-d}$ and $T_\Theta: \mathbb{R}^{d} \rightarrow \mathbb{R}^{D-d}$ stand for scale function and transformation function. For the sake of the numerical stability of cascading multiple flow layers, we follow Moflow~\cite{zang2020moflow} to replace the exponential function for the $S_\Theta$ with the Sigmoid function:
\begin{equation}
\begin{aligned}
    y_{1:d} &= x_{1:d} \\
    y_{d+1:D} &= x_{d+1:D} \odot \mathrm{Sigmoid} (S_\Theta(x_{1:d})) + T_\Theta(x_{1:d}),
\end{aligned}
\end{equation}
and the invertibility is guaranteed by:
\begin{equation}
\begin{aligned}
    x_{1:d} &= y_{1:d} \\
    x_{d+1:D} &= (y_{d+1:D} - T_\Theta(y_{1:d})) / \mathrm{Sigmoid}(S_\Theta(y_{1:d})).
\end{aligned}
\end{equation}
The logarithmic Jacobian determinant is:
\begin{equation}
\begin{aligned}
    \log \big|\mathrm{det}\frac{\partial y}{\partial x}\big| &= \log \bigg|\mathrm{det}(\begin{bmatrix}
        \mathbb{I} & 0  \\
        \frac{\partial y_{d+1:D}}{\partial x_{1:d}} & \mathrm{Sigmoid}(S_\Theta(x_{1:d}))  \\
        \end{bmatrix}) \bigg|
    \\
    &= \log \mathrm{Sigmoid}(S_\Theta(x_{1:d})).
\end{aligned}
\end{equation}
To further improve the invertible mapping with more expressive structures and high numerical stability, Glow~\cite{kingma2018glow} proposes using invertible $1 \times 1$ convolution to learn an optimal partition and actnorm layer to normalize dimensions in each channel over a batch by an affine transformation. Invertible $1 \times 1$ convolution is initialized as a random rotation matrix with zero log-determinant and works as a generalization of a permutation of channels. Act norm initializes the scale and the bias such that the post-actnorm activations per-channel have zero mean and unit variance and learns these parameters in training instead of using batch statistics as batch normalization does. 

\subsection{Flow on the Molecular Graph} Prior works on flow-based molecular graph generation are well developed. Inspired by the graph normalizing flows of GRevNets~\cite{liu2019graph}, GraphNVP~\cite{madhawa2019graphnvp} proposes to generate atom features conditioned on the pre-generated adjacency tensors, which is then followed by other one-shot flow-based molecular graph generation approaches, e.g., GRF~\cite{honda2019graph} and Moflow~\cite{zang2020moflow}. Our proposed SiamFlow follows this manner, that is, firstly transforms the bonds $B$ of molecules to the latent variables $Z_B$ with Glow~\cite{kingma2018glow}, and then transforms the atom features $A$ given $B$ into the conditional latent variable $Z_{A|B}$ with a graph conditional flow.

Let $N, K, C$ be the number of nodes, node types, and edge types, respectively. A molecular graph $G=(A, B)$ is defined by an atom matrix $A \in \{0, 1\}^{N \times K}$ and a bond tensor $B \in \{0, 1\}^{C\times N \times N}$, which correspond to nodes and edges in the vanilla graph. $A[i, k]=1$ represents the $i$-th atom $i$ has atom type $k$, and $B[c, i, j] = 1$ represents there is a bond with type $c$ between the $i$-th atom and $j$-th atom.

Flow-based molecular graph generation methods decompose the generative model into two parts:
\begin{equation}
    P(G) = P((A, B)) \approx P(A | B; \theta_{A|B}) P(B; \theta_{B}),
\end{equation}
where $\theta_{B}$ is learned by the bond flow model $h_B$, and $\theta_{A|B}$ is learned by the atom flow model $h_{A|B}$ conditioned on the bond tensor $B$.

With the strengths of the flow, the optimal parameters $\theta^*_{A|B}$ and $\theta^*_{B}$ maximize the exact likelihood estimation:
\begin{align}
    \arg \max_{\theta_{A|B}, \theta_{B}} \mathbb{E}_{(A, B) \sim P_{G}}[\log P(A|B;\theta_{A|B}) + \log P(B;\theta_B)]
\end{align}
Our work follows the one-shot molecular graph generation manner~\cite{madhawa2019graphnvp,honda2019graph,zang2020moflow} that employs Glow~\cite{kingma2018glow} as the bond flow model $h_B$ and graph conditional flow as the atom flow model $h_{A|B}$.

% \textcolor{red}{TODO: whether add explanations about Glow and graph conditional flow?}

\section{SiamFlow}

\subsection{Overview} 
% 一方面针对现有的分子图生成模型，他们没有conditonal信息
While current flow-based molecular graph generation methods~\cite{madhawa2019graphnvp,honda2019graph,shi2019graphaf,zang2020moflow,kuznetsov2021molgrow,lippe2020categorical,luo2021graphdf} learn from drug-like datasets and generate without the invention of targets, our proposed SiamFlow aims to serve as a conditional flow toward molecular graph generation. 
% 一方面针对现有的conditonal flow模型，他们没有针对图结构做
Though the conditional flow has been well developed in computer vision~\cite{liu2019conditional,kondo2019flow,abdelhamed2019noise,kumar2019videoflow,sun2019dual,pumarola2020c,liang2021hierarchical,yang2021mol2image}, there are limited works that can fit graph generation, especially when it comes to the molecular graph. 

\begin{figure*}[ht]
    \centering
    \includegraphics[width=1.00\textwidth]{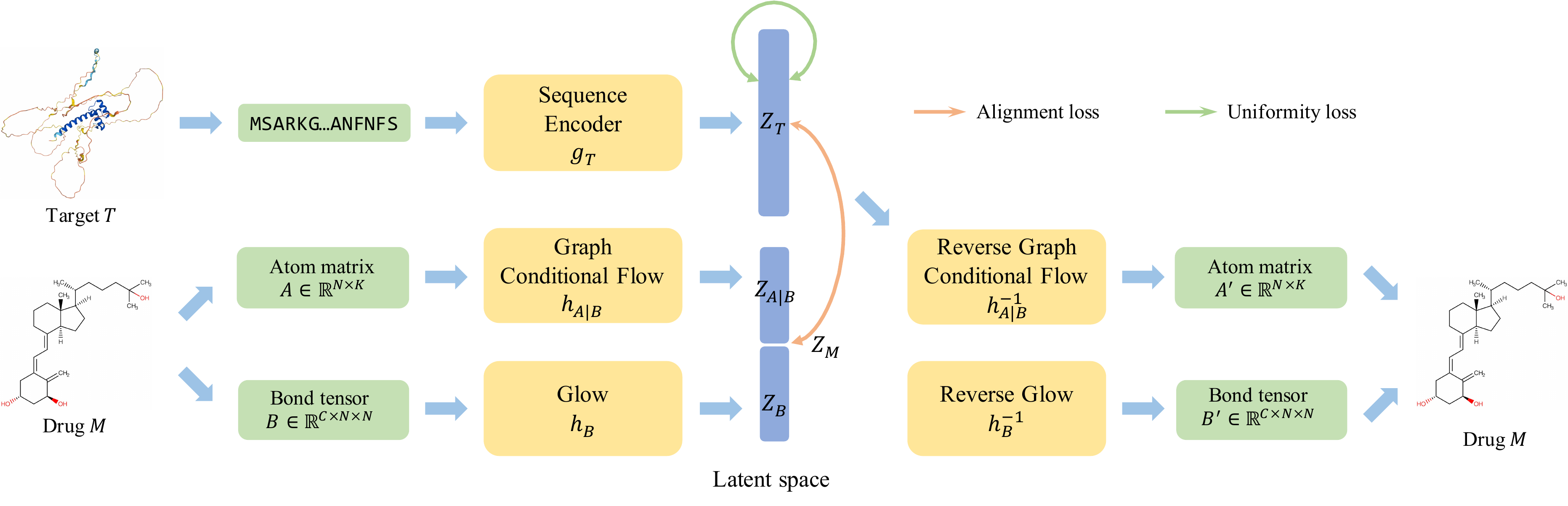}
    \caption{The framework of our proposed SiamFlow. In the training phase, the target sequence embedding $Z_T$ aligns with the drug graph embedding $Z_M$, while a uniformity regularization term forces its distribution as a spherical uniform distribution. In the generation phase, the target sequence embedding $Z_T$ is fed into reverse flows to generate the desired drug.}
    \label{fig:siamflow_framework}
  \end{figure*}

In this section, we introduce SiamFlow, a novel molecular graph generative model conditioned on specific targets. As shown in Figure~\ref{fig:siamflow_framework}, SiamFlow learns the distribution of sequence embedding instead of the isotropic Gaussian distribution like other flow-based methods. 

\subsection{Alignment Loss}

Given a pair of target $T$ and drug $M$, we decompose the drug $M$ into an atom matrix $A \in \mathbb{R}^{N \times K}$ and a bond tensor $B \in \mathbb{R}^{C \times N \times N}$. The sequence encoder $g_T$ can be arbitrary mapping that maps the target sequence $T$ into the sequence embedding $Z_T \in \mathbb{R}^D$. The flow model contains a glow $h_B: \mathbb{R}^{C \times N \times N} \rightarrow \mathbb{R}^{\frac{D}{2}}$ and a graph conditional flow $h_{A|B}: \mathbb{R}^{N \times K} \rightarrow \mathbb{R}^{\frac{D}{2}}$. The drug graph embedding $Z_M$ is the concatenation of $Z_{A|B}$ and $Z_{B}$.

Instead of directly learning the isotropic Gaussian distribution, we impose alignment loss between the target sequence embedding $Z_T$ and the drug graph embedding $Z_M$ so that $Z_T$ can be used as the input of the generation process. Thus, the generated atom matrix and the bond tensor are: 
\begin{equation}
    A' = h^{-1}_{A|B}(Z_T[1:\frac{D}{2}]) ,\; B' = h^{-1}_{B}(Z_T[\frac{D}{2}:D]).
\end{equation}
While traditional flow-based models assume the latent variables follow the Gaussian distribution, SiamFlow forces the flow model to learn the distribution of the condition information instead of a predefined distribution. We define the alignment loss $\mathcal{L}_{align}$ as:
\begin{equation}
\begin{aligned}
    \mathcal{L}_{align} :&= \mathbb{E}_{(T, M)\sim P_{\mathrm{data}}}||Z_T - Z_M||_2 \\ 
    &= \mathbb{E}_{(T, M)\sim P_{\mathrm{data}}}||Z_T - [Z_{A|B}, Z_B]||_2
\end{aligned}
\end{equation}
where $[Z_{A|B}, Z_B]$ denotes the concatenation of the atom embedding $Z_{A|B}$ and the bond embedding $Z_B$, and the pair of protein target $T$ and molecular drug $M$ is sampled from the data $P_{\mathrm{data}}$.

The alignment loss bridges the connections between the target sequence embedding $Z_T$ and the drug graph embedding $Z_M$ in the latent space, but there are still challenges that will be revealed in Sec.~\ref{lab:uniformity_loss} and Sec.~\ref{lab:one-target-to-many-drugs}.

\subsection{Uniformity Loss}
\label{lab:uniformity_loss}

Simply aligning the target sequence embedding $Z_T$ and the drug graph embedding $Z_M$ is not enough. There still remains three challenges: (1) the distribution of $Z_T$ is uncertain, so that the alignment learning may be difficult to converge; (2) sampling from an unknown distribution is indefinite in the generation process; (3) the alignment loss alone admits collapsed solutions, e.g., outputting the same representation for all targets.

\begin{figure}[ht]
    \centering
    \includegraphics[width=0.474\textwidth]{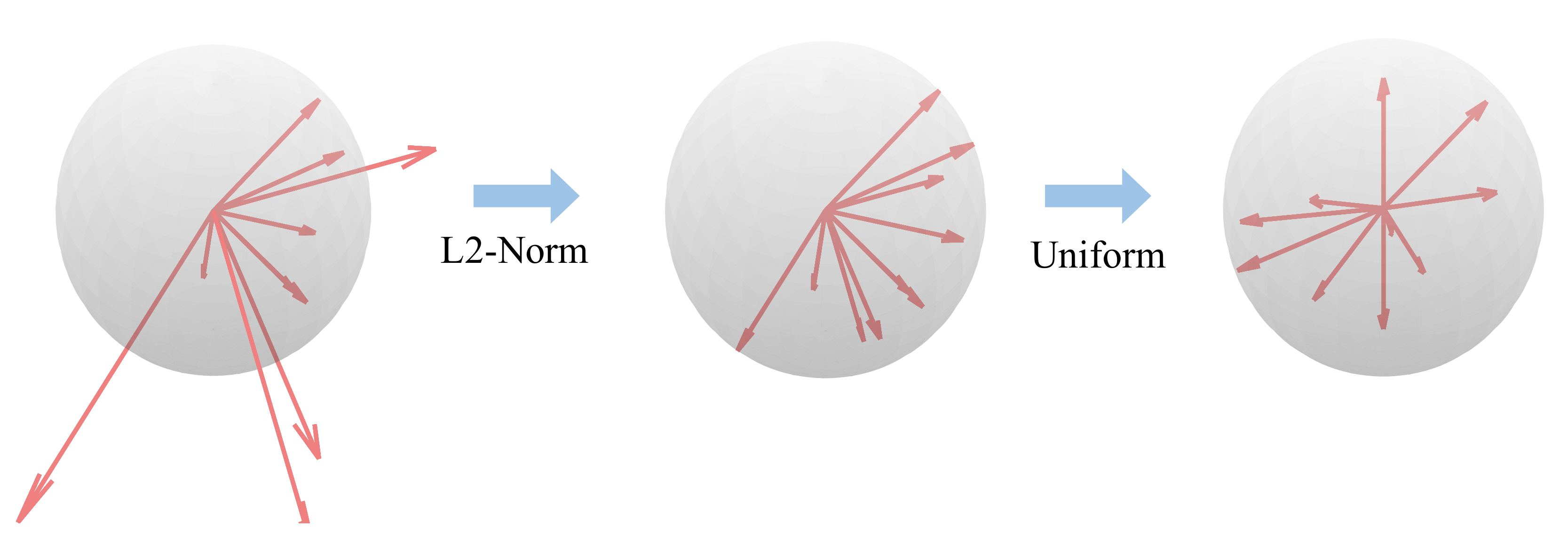}
    \caption{The schematic diagram of the uniformity loss.}
    \label{fig:uniform_loss}
\end{figure}

To overcome the above issues, we design an objective to force the target sequence embedding $Z_T$ to follow a specific distribution, in our case the uniform distribution on the unit hypersphere~\cite{saff1997distributing,kuijlaars1998asymptotics,hardin2004discretizing,liu2018learning,borodachov2019discrete,wang2020understanding}. We recognize angles of embeddings are the critical element that preserves the most abundant and discriminative information. By fitting the hyperspherical uniform distribution, the projections of target sequence embeddings on the hypersphere are kept as far away from each other as possible; thus, discriminations are imposed. Specifically, we project the target sequence embedding $Z_T$ into a unit hypersphere $\mathbb{S}^{D-1}$ by L2 normalization and require the embeddings uniformly distributed on this hypersphere, as shown in Figure~\ref{fig:uniform_loss}.

The uniform hypersphere distribution can be formulated as a minimizing pairwise potential energy problem~\cite{liu2018learning,borodachov2019discrete,wang2020understanding} while higher energy implies less discriminations. Let $\widehat{Z}_T = \frac{Z_T}{\|Z_T\|} \in \mathcal{C}$, and $\mathcal{C}$ is a finite subset of the unit hypersphere $\mathbb{S}^{D-1} \in \mathbb{R}^{D}$. We define the $f$-potential energy~\cite{cohn2007universally} of $\mathcal{C}$ to be:
\begin{equation}
    \sum_{\widehat{Z}_T^{(x)}, \widehat{Z}_T^{(y)} \in \mathcal{C}, x \neq y} f(|\widehat{Z}_T^{(x)} - \widehat{Z}_T^{(y)}|^2).
\end{equation}
where $\widehat{Z}_T^{(x)}$ and $\widehat{Z}_T^{(y)}$ denote different normalized sequence embedding with index $x$ and $y$. 

\noindent{\textbf{Definition.} (Universally optimal~\cite{cohn2007universally}).} A finite subset $\mathcal{C} \subset \mathbb{S}^{D-1}$ is \textit{universally optimal} if it (weakly) minimizes potential energy among all configurations of $|\mathcal{C}|$ points on $\mathbb{S}^{D-1}$ for each completely monotonic potential function.

In SiamFlow, we consider the Gaussian function kernel $G_t(x, y) : \mathbb{S}^{D-1}\times \mathbb{S}^{D-1}\rightarrow \mathbb{R} $ as the potential function $f$, which is defined as:
\begin{equation}
    \label{equ:gaussain_kernl}
    G_t(x, y) = e^{-t |x - y|^2}.
\end{equation}
This kernel function is closely related to the universally optimal configuration, and distributions of points convergence weak* to the uniform distribution by minimizing the expected pairwise potential. 

\noindent{\textbf{Theorem.} (Strictly positive definite kernels on $\mathbb{S}^{D}$~\cite{borodachov2019discrete}).} Consider kernel $K_f: \mathbb{S}^{D} \times \mathbb{S}^{D} \rightarrow(-\infty,+\infty]$ of the form $K_f(x, y):= f(|x-y|^2)$, if $K_f$ is \textit{strictly positive definite} on $\mathbb{S}^{D} \times \mathbb{S}^{D}$ and the energy $I_{K_f}[\sigma_D]$ is finite, then $\sigma_D$ is the unique measure on Borel subsets of $\mathbb{S}^D$ in the solution of $\min_{\mu \in \mathcal{M}(\mathbb{S}^D)} I_{K_f}(\mu)$, and the normalized counting measure associated with any $K_f$-energy minimizing sequence of point configurations on $\mathbb{S}^D$ converges weak* to $\sigma_D$.

This theorem reveals the connections between strictly positive definite kernels and the energy minimizing problem. The Gaussian function is strictly positive definite on $\mathbb{S}^D \times \mathbb{S}^{D}$, thus well tied with the uniform distribution on the unit hypersphere.

\noindent{\textbf{Proposition 1.} (Strictly positive definite of the Gaussian function)} For any $t > 0$, the Gaussian function kernel $G_t(x, y)$ is strictly positive definte on $\mathbb{S}^{D} \times \mathbb{S}^{D}$.

\textit{Proof.} See appendix.

Though Riesz s-kernels $R_s(x, y):= |x-y|^{-s}$ are commonly used as potential functions, we argue that the Gaussian function is expressive because it maps distances to infinite dimensions like radial basis functions, benefiting from the Taylor expansion of exponential functions. Moreover, the Gaussian function is a general case of Riesz s-kernels and can represent Riesz s-kernels by:
\begin{equation}
    R_s(x, y) = \frac{1}{\Gamma(s/2)} \int_{0}^{\infty} G_t(x, y) t^{s/2 - 1} dt.
\end{equation}
where $\Gamma(s/2) = \int_{0}^{\infty} e^{-t} t^{s/2-1}$ for $s > 0$.

As the Gaussian functionk kernel is an ideal choice of potential functions, we define the uniformity loss as the logarithm of the pairwise Gaussian potential's expectation:
\begin{equation}
    \mathcal{L}_{unif}:= \log \mathbb{E}_{(T^{(x)}, T^{(y)})\sim P_{\mathcal{T}}}[G_t(\widehat{Z}_T^{(x)}, \widehat{Z}_T^{(y)})],
\end{equation}
where $T^{(x)}$ and $T^{(y)}$ are two different targets sampled from the target data $P_{\mathcal{T}}$.

\subsection{One Target to Many Drugs}
\label{lab:one-target-to-many-drugs}

\begin{figure}[ht]
    \centering
    \includegraphics[width=0.49\textwidth]{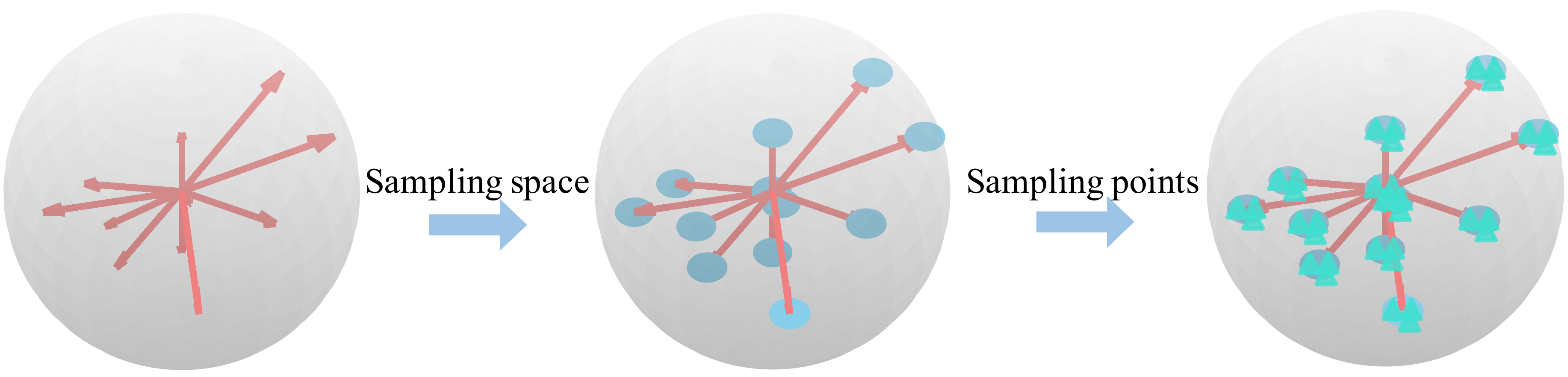}
    \caption{The schematic diagram of the one-to-many strategy. The blue circles denote the possible spaces around the target sequence embeddings, and the green triangles denote the instances sampled from the possible spaces.}
    \label{fig:one-to-many}
\end{figure}

Implementing the alignment loss and the uniformity loss above, the flow model can already generate validated molecular drugs conditioned on specific targets. However, there are multiple affinable drugs for a single target in most cases. To deal with this \textit{one-to-many} problem, we reformulate learning target embeddings into learning  spaces of target embeddings in the latent space, as shown in Figure~\ref{fig:one-to-many}. 

As the target embeddings have been pushed by the uniformity loss to stay as far away as possible on the hypersphere, they preserve abundant and discriminative information to a large extent. We design an adaptive space learning strategy that holds the discriminative angle information with a limited scope. For a set of target sequence embeddings $\mathcal{Z}_\mathcal{T} = \{Z_T^{(0)}, ..., Z_T^{(L)}\}$, we first calculate their standard deviation by:
\begin{equation}
    \sigma(\mathcal{Z}_{\mathcal{T}}) = \sqrt{\frac{1}{L} \sum_{i=1}^L (Z_T^{(i)} - \mu(\mathcal{Z}_{\mathcal{T}}))},
\end{equation}
where $\mu(\mathcal{Z}_{\mathcal{T}}) = \frac{1}{L} \sum_{i=1}^L Z_T^{(i)}$ is the mean of the set $\mathcal{Z}_{\mathcal{T}}$. 
Then, we define a space for each target sequence embedding:
\begin{equation}
    \Omega(Z_T) = \{Z_T + Z'_T| Z'_T \in \mathcal{N}(0, \lambda \sigma^2(\mathcal{Z}_\mathcal{T}))\},
\end{equation}
where $\lambda$ is the hyperparameter that controls the scale of the space and is empirically set as 0.1. 

Note that we define the space on $Z_T$ instead of the normalized $\widehat{Z}_T$, as normalized embeddings lose the length information to the extent that the available space is limited. 

Thus, we modify the alignment loss as:
\begin{equation}
    \mathcal{L}_{align} = \mathbb{E}_{(T, M)\sim P_{\mathrm{data}}}|\Omega(Z_T) - Z_M|.
\end{equation}
In the generation process, sampling from the same space is permissible to generate desired drugs.

In summary, the objective is a linear combination of the modified alignment loss and uniform loss:
\begin{equation}
    \mathcal{L}_{total} = \mathcal{L}_{align} + \mathcal{L}_{unif}
\end{equation}
% 需要整体算法流程或者伪代码嘛？

\section{Experiments}

\paragraph{Baselines} 
Since we present a novel generative approach conditioned on targets, we primarily compare our approach to other conditional generative models, i.e., conditional VAE~\cite{sohn2015learning}. Furthermore, an attention-based Seq2seq~\cite{sutskever2014sequence,vaswani2017attention} neural translation model between the target protein sequence and drug SMILES is considered a straightforward solution in our setting. An explainable substructure partition fingerprint~\cite{huang2019explainable} is employed for sequential drug SMILES and protein sequences. We unify these baselines and our proposed SiamFlow into a unified framework for fair comparisons. 

\paragraph{Datasets} To evaluate the ability of our proposed SiamFlow, we collect a dataset based on four drug-target interaction datasets, including BIOSNAP~\cite{zitnik2018biosnap}, BindingDB~\cite{liu2007bindingdb}, DAVIS~\cite{davis2011comprehensive}, and DrugBank~\cite{wishart2018drugbank}. We remove all the negative samples in the original datasets, and only keep the positive samples. Our dataset contains 24,669 unique drug-target pairs with 10,539 molecular drugs and 2,766 proteins. The maximum number of atoms in a molecular drug is 100 while 11 types of common atoms are considered. We define the dataloader to ensure zero overlap protein in the training, validation, and test set.

\paragraph{Metrics}
% 1. for generative models
% 2. chemical 
To comprehensively evaluate the conditional generative models in terms of target-aware molecular generation, we design metrics from two perspectives: (1) Generative metrics. Following the common molecular generation settings, we apply metrics including: \textbf{validity} which is the percentage of chemically valid molecules in all the generated molecules, \textbf{uniqueness} which is the percentage of unique valid molecules in all the generated molecules, \textbf{novelty} which is the percentage of generated valid molecules which are not in the training dataset. (2) Chemical metrics. We evaluate the similarities between the generated drugs and the nearest drugs in the training set including: \textbf{Tanimoto similarity} which is calculated based on hashed binary features,
\textbf{Fraggle similarity} which focus on the fragment-level similarity, \textbf{MACCS similarity} which employs 166-bit 2D structure fingerprints.

\paragraph{Empirical Running Time}
% This paragraph refers Moflow
We implement our proposed method SiamFlow and the other two baselines Seq2seq, CVAE by Pytorch-1.8.1 framework. We train them with Adam optimizer with learning rate 0.001, batch size 16, and 100 epochs on a single NVIDIA Tesla V100 GPU. To evaluate the validity and chemical similarities, we employ the cheminformatics toolkit RDKit in the assessment phase. Our SiamFlow completes the training process of 100 epochs in an average of 1.06 hours (38 seconds/ epoch), while CVAE and Seq2seq take an average of 1.14 hours (41 seconds/ epoch) and 8.33 hours (5 minutes/ epoch) respectively.

\subsection{Target-aware Molecular Graph Generation}

We conduct experiments on molecular drug generation with specific targets for comparisons. For each experiment, we repeat three trials with different random seeds and report the mean and standard deviation. 

Table~\ref{tab:results_generative} shows the results on generative metrics of our SiamFlow model in comparison to the baselines. Our proposed SiamFlow inherits the strengths of the flow and far surpasses other baselines in generative metrics. It can be seen that Seq2seq suffers from low validity, uniqueness, and novelty, which indicates Seq2seq's generation relies on its memorization. CVAE has higher uniqueness and novelty than Seq2seq though its validity is even lower. Besides, the standard deviations of metrics on CVAE are relatively high, suggesting it is volatile to train. Among them, SiamFlow obtains superior performance with the least volatility.

\begin{table}[ht]
    \vskip -0.10in
    \caption{Evaluation results on generative metrics of SiamFlow v.s. baselines; high is better for all three metrics here. }
    \label{tab:results_generative}
    \begin{center}
    \setlength{\tabcolsep}{1.5mm}{
    \begin{tabular}{lcccr}
    \toprule
    Method & \% Validity & \% Uniqueness & \% Novelty \\
    \midrule
    Seq2seq    & 16.08$\pm$4.14 & 13.87$\pm$1.74 & 14.89$\pm$11.41 \\
    CVAE       & 12.54$\pm$7.56 & 72.30$\pm$20.33 & 99.72$\pm$0.39 \\
    SiamFlow   & \textbf{100.00}$\pm$0.00 & \textbf{99.61}$\pm$0.16 & \textbf{100.00}$\pm$0.00 \\
    \bottomrule
    \end{tabular}}
    \end{center}
    \vskip -0.10in
\end{table}

In addition to generative metrics, we also report chemical metrics in Table~\ref{tab:results_chemical}. The generated molecular drugs are expected to have a chemical structure similar to the ground-truth drugs in order to have a high binding affinity to the target. SiamFlow is consistently better than other baselines in both the Tanimoto and Fraggle similarity while obtaining relatively lower MACCS similarity than Seq2seq. Considering that MACCS measures the similarity of encodings of molecules, the sequence partition rules of Seq2seq may help it. Thus, we pay more attention to the Tanimoto and Fraggle similarity because they are structure-centric metrics.

\begin{table}[ht]
    \vskip -0.10in
    \caption{Evaluation results on chemical metrics of SiamFlow v.s. baselines; high is better for all three metrics here. }
    \label{tab:results_chemical}
    \begin{center}
    \setlength{\tabcolsep}{2.3mm}{
    \begin{tabular}{lcccr}
    \toprule
    Method & \% Tanimoto & \% Fraggle & \% MACCS \\
    \midrule
    Seq2seq    & 26.27$\pm$9.91 & 25.84$\pm$7.27 & \textbf{37.98}$\pm$7.70 \\
    CVAE       & 7.76$\pm$6.61 & 12.31$\pm$5.81 & 16.42$\pm$7.17 \\
    SiamFlow   & \textbf{48.55}$\pm$0.97 & \textbf{34.41}$\pm$0.35 & 29.30$\pm$1.07 \\
    \bottomrule
    \end{tabular}}
    \end{center}
    \vskip -0.10in
\end{table}

\begin{figure}[ht]
    \centering
    \includegraphics[width=0.48\textwidth]{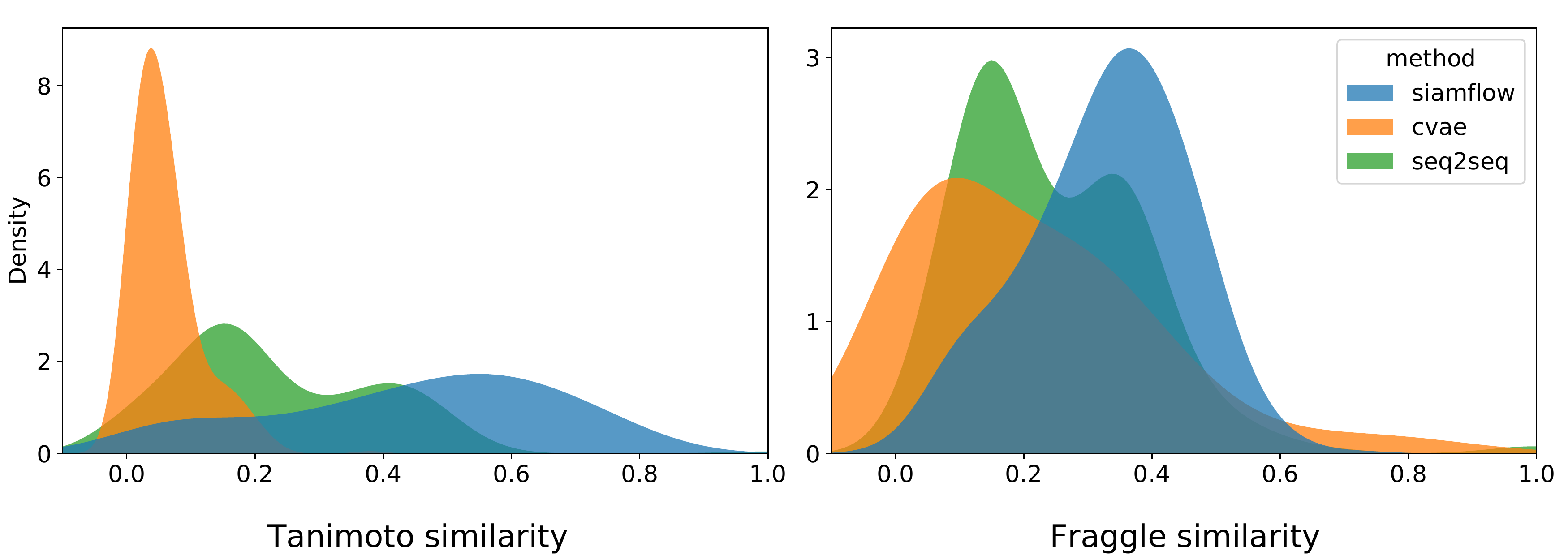}
    \caption{The distribution of generative metrics evaluated on SiamFlow and other baselines.}
    \label{fig:density}
\end{figure}

\begin{figure}[ht]
    \centering
    \includegraphics[width=0.45\textwidth]{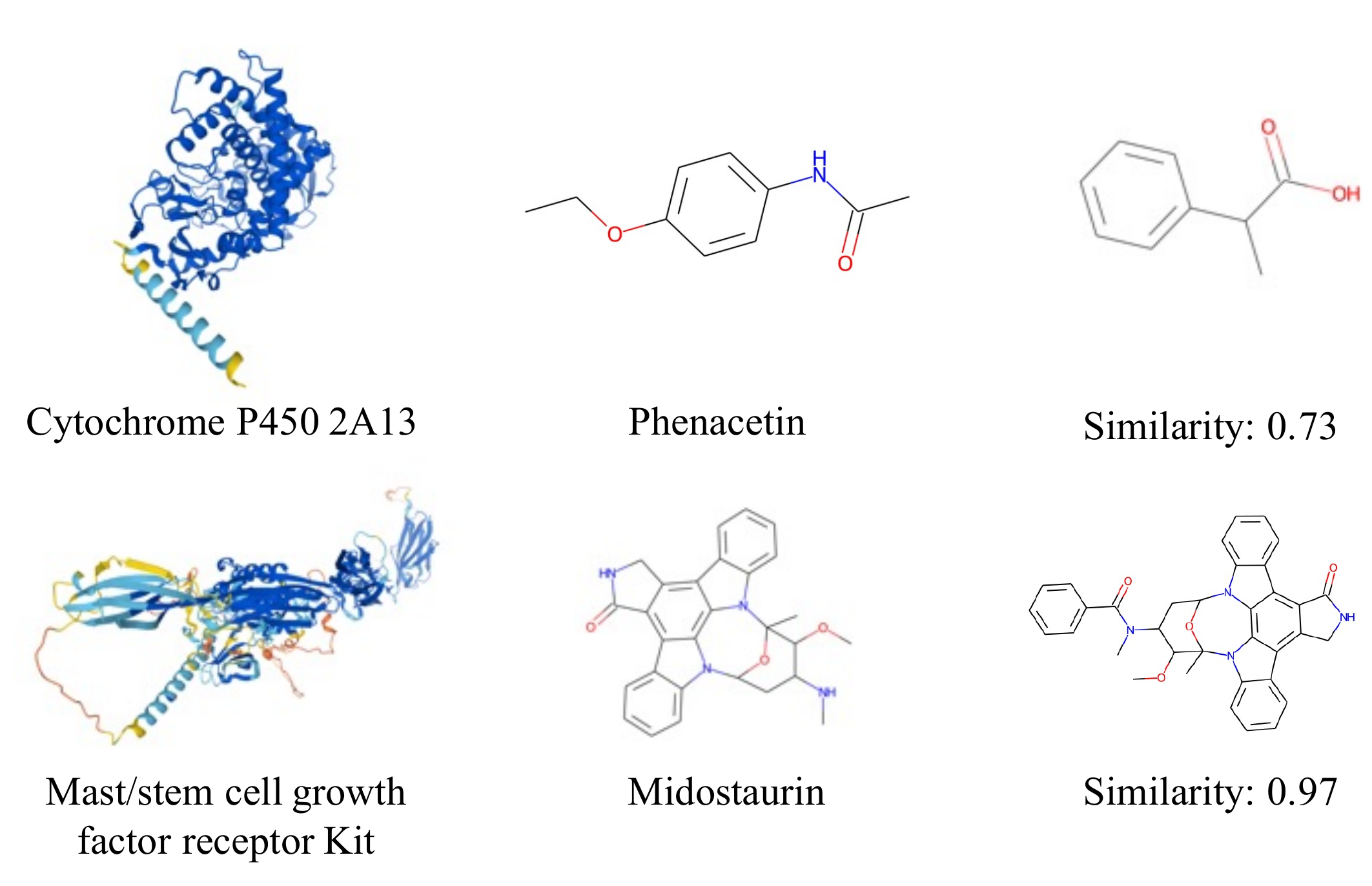}
    \caption{Examples of the generated drugs.}
    \label{fig:example}
\end{figure}

We visualize the distribution of the Tanimoto similarity and the Fraggle similarity evaluated on these methods in Figure~\ref{fig:density}. SiamFlow consistently outperforms other methods and generates desirable molecular drugs. The examples of generated drugs are shown in Figure~\ref{fig:example}.

\subsection{Ablation Study}

We conduct the ablation study and report the results in Table~\ref{tab:ablation_generative} and Table~\ref{tab:ablation_chemical}. It can be seen from Table~\ref{tab:ablation_generative} that simply aligning the target sequence embedding and drug graph embedding will result in extremely low uniqueness. Our one-to-many strategy enriches the latent space so that one target can map to different drugs. The absence of $\mathcal{L}_{unif}$ does not harm the generative metrics because it only constrains the distribution of target sequence embeddings but has a limited impact on the generation process.

\begin{table}[ht]
    \vskip -0.10in
    \caption{Ablation results on generative metrics.}
    \label{tab:ablation_generative}
    \begin{center}
    \setlength{\tabcolsep}{1.2mm}{
    \begin{tabular}{lcccr}
    \toprule
    Method & \% Validity & \% Uniqueness & \% Novelty \\
    \midrule
    SiamFlow                   & 100.00 & 99.39 & 100.00 \\
    \hline
    w/o one-to-many            & 100.00 & 12.55 & 100.00 \\
    w/o $\mathcal{L}_{unif}$   & 100.00 & 100.00 & 100.00 \\
    \bottomrule
    \end{tabular}}
    \end{center}
    \vskip -0.10in
\end{table}

Table~\ref{tab:ablation_chemical} demonstrates the chemical metrics are well without the one-to-many strategy. If we generate only one drug for a particular target, the nearest drug similarity degrades to a special case, i.e., comparing the generated drug with its corresponding one in the training set. Moreover, removing $\mathcal{L}_{unif}$ severely impairs the chemical performance, suggesting the uniformity loss promotes the expressive abilities of target sequence embeddings. 

\begin{table}[ht]
    \vskip -0.10in
    \caption{Ablation results on chemical metrics.}
    \label{tab:ablation_chemical}
    \begin{center}
    \setlength{\tabcolsep}{1.4mm}{
    \begin{tabular}{lcccr}
    \toprule
    Method & \% Tanimoto & \% Fraggle & \% MACCS \\
    \midrule
    SiamFlow                   & 49.43 & 34.62 & 29.55 \\
    \hline
    w/o one-to-many            & 48.83 & 34.93 & 31.23 \\
    w/o $\mathcal{L}_{unif}$   & 18.49 & 15.70 & 17.91 \\
    \bottomrule
    \end{tabular}}
    \end{center}
    \vskip -0.10in
\end{table}

\section{Conclusion and Discussion}

In this paper, we explore target-aware molecular graph generation, which differs in generating drugs conditioned on specific targets, while existing methods focus on developing similar drugs based on drug-like datasets. Target-aware molecular generation combines drug-like molecular generation with target-specific screening and thus simplifies the drug-target interaction step. 

To explore this problem in-depth, we compile a benchmark dataset from several public datasets and build baselines in a unified framework. Furthermore, we take advantage of recent progress on flow-based molecular graph generation methods and propose SiamFlow towards target-aware molecular generation. With the alignment and uniform loss, the proposed method can effectively generate molecular drugs conditioned on protein targets. Moreover, we deal with one target to many drugs by aligning the embedding space instead of a single embedding. 

Extensive experiments and analyses demonstrate that SiamFlow is a promising solution towards target-aware molecular generation. A worthwhile avenue for future work is introducing protein structure information for decent target sequence representations.

\newpage

% \section*{Acknowledgements}


\bibliography{ref}
\bibliographystyle{icml2021}

\end{document}